\documentclass[11pt,a4paper]{article}
\usepackage[hyperref]{emnlp2020}
\usepackage{times}
\usepackage{latexsym}

\usepackage{microtype}

\aclfinalcopy 
\def\aclpaperid{593} 

\usepackage{subcaption}

\usepackage{graphicx}
\graphicspath{ {figures/} }
\usepackage[T1]{fontenc} 
\usepackage{multirow}
\usepackage{float} 
\usepackage{booktabs} 

\usepackage[utf8]{inputenc}
\usepackage{pgfplots}
\usepackage{layouts}
\usepackage{tikz}

\title{The Grammar of Emergent Languages\\
\vspace{3mm}}

\author{Oskar van der Wal\\
University of Amsterdam\\
  \texttt{oskar.vanderwal@gmail.com} \\\And
  Silvan de Boer\\
  University of Amsterdam\\
  \texttt{silvandeboer@gmail.com}\\
  \\\AND
  Elia Bruni$^*$ \\
  Institute of Cognitive Science\\
  University of Osnabruek\\
  {\tt elia.bruni@gmail.com} \\
  \\\And
  Dieuwke Hupkes\thanks{~ Shared senior authorship} \\
  Institute for Logic, Language and Computation\\
  University of Amsterdam \\
  {\tt d.hupkes@uva.nl}
}

\date{}

\begin{document}
\maketitle

\begin{abstract}
In this paper, we consider the syntactic properties of languages emerged in referential games, using unsupervised grammar induction (UGI) techniques originally designed to analyse natural language.
We show that the considered UGI techniques are appropriate to analyse emergent languages and we then study if the languages that emerge in a typical referential game setup exhibit syntactic structure, and to what extent this depends on the maximum message length and number of symbols that the agents are allowed to use.
Our experiments demonstrate that a certain message length and vocabulary size are required for structure to emerge, but they also illustrate that more sophisticated game scenarios are required to obtain syntactic properties more akin to those observed in human language.
We argue that UGI techniques should be part of the standard toolkit for analysing emergent languages and release a comprehensive library to facilitate such analysis for future researchers.

\end{abstract}

\section{Introduction}
Artificial agents parameterised by deep neural networks can learn to communicate using discrete symbols to solve collaborative tasks \citep[][]{foerster2016learning,lazaridouMultiAgentCooperationEmergence2017,havrylov2017emergence}. %
A prime reason to conduct such studies, which constitute a new generation of experiments with referential games, is that they may provide insight in the factors that shaped the evolution of human languages \citep{kirby2002natural}.

However, the \emph{emergent languages} developed by neural agents are not human-interpretable, and little is known about their semantic and syntactic nature.
More specifically, we do not know to what extent the structure of emergent languages resembles the structure of human languages, what the languages encode, and how these two things depend on choices that need to be made by the modeller. %

A substantial obstacle to better understanding emergent languages is the lack of tools to analyse their properties. %
Previous work has concentrated primarily on understanding languages through their \emph{semantics}, by studying the alignment of messages and symbolic representations of the meaning space \citep[e.g.][]{lazaridouEmergenceLinguisticCommunication2018}.
A substantial downside of such approaches is that they are restricted to scenarios for which a symbolic representation of the meaning space is available.
Furthermore, they ignore a second important aspect of language: syntax, which is relevant not just for syntactically-oriented researchers, but also for those that are interested in semantics from a compositional perspective.
In this work, we aim to address this gap in the literature by presenting an analysis of the syntax of emergent languages.

We take inspiration from unsupervised grammar induction (UGI) techniques originally proposed for natural language. %
In particular, we use them to investigate if the languages that emerge in the typical setup of referential games exhibit interesting syntactic structure, and to what extent this depends on the maximum message length and number of symbols that the agents are allowed to use.

We first establish that UGI techniques are suitable also for our artificial scenario, by testing them on several artificial structured languages that are distributionally similar to our emergent languages.
We then use them to analyse a variety of languages emerging from a typical referential game, with various message lengths and vocabulary sizes.
We show that short messages of up to length five do not give rise to any interesting structure, while longer messages are significantly more structured than \emph{random} languages, but yet far away from the type of syntactic structure observed in even simple human language sentences.

As such, our results thus suggest that more interesting games scenarios may be required to trigger properties more similar to human syntax and -- importantly -- confirm that UGI techniques are a useful tool to analyse such more complex scenarios.
Their results are informative not only for those interested in the evolution of \emph{structure} of human languages, but can also fuel further semantic analysis of emergent languages.

\section{Related work}

Previous work that focused on the analysis of emergent languages has primarily concentrated on semantics-based analysis.
In particular, they considered whether agents transmit information about categories or objects, or instead communicate using low-level feature information \cite[i.a.]{steels2010artificial,lazaridouMultiAgentCooperationEmergence2017,bouchacourtHowAgentsSee2018,lazaridouEmergenceLinguisticCommunication2018,mihai2019avoiding}.

\subsection{Qualitative inspection}
Many previous studies have relied on qualitative, manual inspection.
For instance, \citet{lazaridouEmergenceLinguisticCommunication2018} and \citet{havrylov2017emergence} showed that emergent languages can encode category-specific information through prefixing as well as word-order and hierarchical coding, respectively.
Others instead have used qualitative inspection to support the claim that messages focus on pixel information instead of concepts \cite{bouchacourtHowAgentsSee2018}, that agents consistently use certain words for specific situations \cite{mul2019mastering} or re-use the same words for different property values \cite{lu2020countering}, or that languages represent distinct properties of the objects (e.g.\ colour and shape) under specific circumstances \cite{kottur2017natural,choi2018compositional,slowik2020exploring}.

\subsection{RSA} Another popular approach to analyse the semantics of emergent languages relies on \emph{representational similarity analysis} \citep[RSA,][]{kriegeskorte2008representational}.
RSA is used to analyse the similarity between the language space and the meaning space, in which case it is also called \emph{topographic similarity} \citep{brighton2005language,brighton2006understanding,lazaridouEmergenceLinguisticCommunication2018,andreas2019measuring,li2019ease,keresztury2020compositional,slowik2020exploring,ren2020compositional}, %
It has also been used to directly compare the continuous hidden representations of a neural agent with the input space \citep{bouchacourtHowAgentsSee2018}.

\subsection{Diagnostic Classification} A last technique used to analyse emergent languages is \emph{diagnostic classification} \cite{hupkes2018visualisation}, which is used to examine which concepts are captured by the visual representations of the playing agents \cite{lazaridouEmergenceLinguisticCommunication2018}, whether the agents communicate their hidden states \cite{cao2018emergent}, which input properties are best retained by the agent's messages \cite{hupkes2020internal} and whether the agents communicate about their own objects and possibly ask questions \cite{bouchacourt2019miss}.

\section{Method}
\label{sec:setup}

We analyse the syntactic structure of languages emerging in referential games with UGI techniques.
In this section, we describe the game setup that we consider (\S\ref{subsec:game}), the resulting languages that are the subject of our analysis (\S\ref{subsec:languages}) and the UGI techniques that we use (\S\ref{subsec:ugi}).
Lastly, we discuss our main methods of evaluating our UGI setups and the resulting grammars (\S\ref{subsec:evaluation}) as well as several baselines that we use for comparison (\S\ref{subsec:baselines}).

\subsection{Game}\label{subsec:game}
We consider a game setup similar to the one presented by \citet{havrylov2017emergence}, in which we vary the message length and vocabulary size.
In this game, two agents develop a language in which they speak about $30\times30$ pixel images that represent objects of different shapes, colours and sizes ($3 \times 3 \times 2$), placed in different locations.
In the first step of the game, the \emph{sender} agent observes an image and produces a discrete message to describe it.
The \emph{receiver} agent then uses this message to select an image from a set containing the correct image and three distractor images.
Following \citet{hupkes2020internal}, we generate the target and distractor images from a symbolic description with a degree of non-determinism, resulting in $75k$, $8k$, and $40k$ samples for the train, validation, and test set.

Both the sender and receiver agent are modelled by an LSTM and CNN as language and visual units, respectively.
We pretrain the visual unit of the agents by playing the game once, after which it is kept fixed throughout all experiment.
All trained agents thus have the same visual unit, during training only the LSTM's parameters are updated.
We use Gumbel-Softmax with a temperature of $1.2$ for optimising the agents' parameters, with batch size $128$ and initial learning rate $0.0001$ for the Adam optimiser \cite{kingma2015adam}.
In addition to that, we use early stopping with a patience of $30$ to avoid overfitting.
We refer to Appendix \ref{sec:appendix-ref_game} for more details about the architectures and a mathematical definition of the game that we used.

\subsection{Languages}\label{subsec:languages}
From the described game, we obtain several different languages by varying the maximum message length $L$ and vocabulary size $V$ throughout experiments.
For each combination of $L \in \{3,5,10\}$ and $V \in \{6,13,27\}$, we train the agents three times.
In all these runs, the agents develop successful communication protocols, as indicated by their high test accuracies (between $0.95$ and $1.0$).
Furthermore, all agents can generalise to unseen scenarios.

For our analysis, we then extract the sender messages for all $40K$ images from the game's test set.
From this set of messages, we construct a disjoint induction set ($90\%$) and validation set ($10\%$).
Because the sender may use the same messages for several different input images, messages can occur multiple times.
In our experiments, we consider only the set of unique messages, which us this smaller than the total number of images.
Table~\ref{tab:languages} provides an overview of the number of messages in the induction and evaluation set for each language with maximum message length $L$ and vocabulary size $V$.

In the rest of this paper we refer to the three sets by denoting the message length and vocabulary size of the game they come from.
For instance, $V6L10$ refers to the set of languages trained with a vocabulary size of $6$ and a maximum message length of $10$.
Note that while the sender agent of the game may choose to use shorter messages and fewer symbols than these limits, they typically do not.

\begin{table}[!h]
  \setlength\tabcolsep{4pt}
  \centering
  {  \small
  \begin{tabular}{ll|rrrrrr}
    \toprule
    {} & {} & \multicolumn{2}{c}{seed 0} & \multicolumn{2}{c}{seed 1} & \multicolumn{2}{c}{seed 2} \\
    L  & V  & induct. & eval. & induct. & eval. & induct. & eval.            \\
    \midrule
3	&	6	&	162	&	19	&	141	&	16	&	147	&	17	\\
	&	13	&	440	&	49	&	390	&	44	&	358	&	40	\\
	&	27	&	596	&	67	&	554	&	62	&	512	&	57	\\
5	&	6	&	913	&	102	&	795	&	89	&	781	&	87	\\
	&	13	&	1819	&	203	&	1337	&	149	&	1614	&	180	\\
	&	27	&	2062	&	230	&	1962	&	219	&	1429	&	159	\\
10	&	6	&	4526	&	503	&	4785	&	532	&	4266	&	475	\\
	&	13	&	8248	&	917	&	9089	&	1010	&	7546	&	839	\\
	&	27	&	9538	&	1060	&	8308	&	924	&	9112	&	1013	\\
    \bottomrule
  \end{tabular}
}
\caption{The number of messages per language for the induction and evaluation set, for all three seeds for playing the referential game.}
  \label{tab:languages}
\end{table}

\subsection{Grammar induction}\label{subsec:ugi}
\begin{figure}
    \vspace{3mm}
  \centering
{\small
  \usetikzlibrary{shapes.geometric, arrows}
\tikzstyle{startstop} = [rectangle, rounded corners, minimum width=0.5cm, minimum height=0.5cm,text centered, draw=black, fill=red!30]
\tikzstyle{parser} = [trapezium, trapezium left angle=70, trapezium right angle=110, minimum width=0.5cm, minimum height=0.5cm, text centered, draw=black, fill=blue!30]
\tikzstyle{process} = [rectangle, minimum width=3cm, minimum height=1cm, text centered, draw=black, fill=orange!30]
\tikzstyle{decision} = [diamond, minimum width=3cm, minimum height=1cm, text centered, draw=black, fill=green!30]
\tikzstyle{arrow} = [thick,->,>=stealth]

\tikzstyle{cloud} = [draw, circle,
    minimum height=2.5em]
\tikzstyle{block} = [rectangle, draw, 
    text width=3.5em, text centered, rounded corners, minimum height=2em, node distance=2cm]

\begin{tikzpicture}[node distance=1.5cm]
\node (grammar) [cloud] {$G$};
\node (const) [block, right of=grammar] {CCL or DIORA};
\node (bmm) [block, right of=const] {BMM};
\node (induced_grammar) [cloud, right of=bmm] {$G'$};

\draw [arrow] (grammar) -- node[anchor=south] {$\{m\}$} (const);
\draw [arrow] (const) -- (bmm);
\draw [arrow] (bmm) -- (induced_grammar);

\end{tikzpicture}
}
\caption{Our two-stage grammar induction setup. We try to reconstruct the grammar $G$ that is hypothesised to have generated our set of messages $M$, using first CCL and DIORA to infer unlabeled constituency trees for all $m\in M$ and then BMM to label these trees.}
\label{fig:pipeline}
\end{figure}
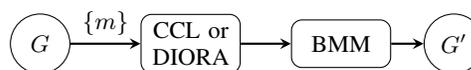

For natural language, there are several approaches to unsupervised parsing and grammar induction.
Some of these approaches induce the syntactic structure (in the form of a bracketing) and the constituent labels simultaneously, but most do only one of those.
We follow this common practice and use a two-stage induction process (see Figure~\ref{fig:pipeline}), in which we first infer unlabelled constituency structures and then label them.
From these labelled structures, we then read out a probabilistic context free grammar (PCFG).

\subsubsection{Constituency structure induction}
To induce constituency structures, we compare two different techniques: the pre-neural statistical \emph{common cover link} parser \citep[CCL,][]{seginer2007fast} and the neural parser \emph{Deep Inside-Outside Recursive Auto-encoder} \citep[DIORA,][]{drozdov2019unsupervised}.\footnote{Another recent and state-of-the-art unsupervised neural parser is the \emph{Unsupervised Recurrent Neural Network Grammar} \citep[URNNG][]{kim2019unsupervised}. For our languages, URNNG generated exclusively right-branching trees, which is why we disregarded it in an initial stage of our experiments.}

\paragraph{CCL}
While proposed in 2007, CCL\footnote{\url{http://www.seggu.net/ccl/}} is still considered a state-of-the-art unsupervised parser.
Contrary to other popular parsers from the 2000s \citep[e.g.][]{klein2004corpus,klein2005natural,ponvert2011simple,reichart2010improved}, it does not require POS-annotation of the words in the corpus, making it appropriate for our setup.

CCL is an incremental and greedy parser, that aims to incrementally add \emph{cover links} to all words in a sentence. 
From these sets of cover links, constituency trees can be constructed.
To limit the search space, CCL incorporates a few assumptions based on knowledge about natural language, such as the fact that constituency trees are generally skewed and the word distribution zipfian.
In our experiments, we use the default settings for CCL.

\paragraph{DIORA}
In addition to CCL, we also experiment with the more recent \emph{neural} unsupervised parser DIORA\footnote{\url{https://github.com/iesl/diora}}.
As the name suggests, DIORA is built on the application of recursive auto-encoders.

In our experiments with DIORA, we use a \textit{tree-LSTM} with a hidden dimension of $50$, and train for a maximum of $5$ epochs with a batch size of $128$.
We use the GloVe framework\footnote{\url{https://nlp.stanford.edu/software/GloVe-1.2.zip}} \citep{pennington2014glove} to pretrain word-embeddings for our corpus; using an embedding size of $16$.

\subsubsection{Constituency labelling}

To label the constituency structures returned by CCL and DIORA, we use \emph{Bayesian Model Merging} \citep[BMM,][]{stolcke1994inducing}.
BMM was originally approached to induce \emph{grammars} for natural language corpora, but proved to be infeasible for that purpose.
However, BMM has been successfully used to infer labels for unlabelled constituency trees \cite{borensztajn2007bayesian}. 
It can therefore complement techniques such as CCL and DIORA.

The BMM algorithm starts from a set of constituency trees in which each constituent is given its own unique label.
It defines an iterative search procedure that merges labels to reduce the joint description length of the data (DDL) and the grammar that can be inferred from the labelling (GDL).
To find the next best merge step, the algorithm computes the effect of merging two labels on the sum of the GDL and DDL after doing the merge, where the GDL is defined as the number of bits to encode the grammar that can be inferred from the current labelled treebank with relative frequency estimation, and the DDL as the negative log-likelihood of the corpus given this grammar.
To facilitate the search and avoid local minima, several heuristics and a look-ahead procedure are used to improve the performance of the algorithm.
We use the BMM implementation provided by \citet{borensztajn2007bayesian}\footnote{\url{https://github.com/pld/BMM_labels/}}.

We refer to our complete setups with the names CCL-BMM and DIORA-BMM, respectively, depending on which constituency inducer was used in the first step.

\subsection{Evaluation}\label{subsec:evaluation}

As we do not know the true structure of the emergent languages, we have to resort to different measures than the traditional \emph{precision}, \emph{recall} and \emph{F1 scores} that are typically used to evaluate parses and grammars.
We consider three different aspects, which we explain below.

\subsubsection{Grammar aptitude} To quantitatively measure how well the grammar describes the data, we compute its \emph{coverage} on a disjoint evaluation set.
Coverage is defined as the ratio of messages that the grammar can parse and thus indicates how well a grammar generalises to unseen messages of the same language.
We also provide an estimate of how many messages \emph{outside} of the language the grammar can parse -- i.e.\ to what extent the grammar \emph{overgenerates} -- by computing its coverage on a subset of 500 randomly sampled messages.

\subsubsection{Language compressibility}
To evaluate the extent to which the grammar can compress a language, we consider the grammar and data description lengths (\emph{GDL} and \emph{DDL}), as defined by \citet{borensztajn2007bayesian}.
To allow comparison between languages that have a different number of messages, we consider the average message DDL.%

\subsubsection{Grammar nature}
Lastly, to get a more qualitative perspective in the nature of the induced grammar, we consider a few statistics expressing the number of \emph{non-terminals} and \emph{pre-terminals} in the grammar, as well as the number of \emph{recursive production rules}, defined as a production rule where the symbol from the left-hand side also appears on the right-hand side.
Additionally, we consider the distribution of \emph{depths} of the most probable parses of all messages in the evaluation sets.

\subsection{Baselines}\label{subsec:baselines}
To ground our interpretation, we compare our induced grammars with three different  language baselines that express different levels of structure.
We provide a basic description here, more details can be found in Appendix~\ref{subsec:appendix-baselines}.

\subsubsection{Random baseline}
We compare all induced grammars with a grammar induced on a random language that has the same vocabulary and length distribution as the original language, but whose messages are sampled completely randomly from the vocabulary.

\subsubsection{Shuffled baseline}
We also compare the induced grammars with a grammar induced on languages that are constructed by \emph{shuffling} the symbols of the emergent languages.
The symbol distribution in these languages are thus identical to the symbol distribution in the languages they are created from, but the symbol \emph{order} is entirely random.

\subsubsection{Structured baseline}
Aside from (semi)random baselines, we also consider a \emph{structured baseline}, consisting of a grammar induced on languages that are similar in length and vocabulary size, but that are generated from a context-free grammar defining a basic hierarchy and terminal-class structure.\footnote{A full description, including some example grammars, can be found in Appendix~\ref{sec:appendix-struct_lang}.}
These structured baseline grammars indicate what we should expect if a relatively simple but yet hierarchical grammar would explain the emergent languages.

\section{Suitability of induction techniques}

As the grammar induction techniques we apply are defined for natural language, they are not trivially also suitable for emergent languages.
In our first series of experiments, we therefore assess the suitability of the grammar induction techniques for our artificial scenario, evaluate to what extent the techniques are dependent on the exact sample taken from the training set, and we determine what is a suitable data set size for the induction techniques.
The findings of these experiments inform and validate the setup for analysing the emergent languages in \S\ref{sec:experiment2}.

\subsection{Grammars for structured baselines}\label{subsec:struct_grammars}
We first qualitatively assess the extent to which CCL-BMM and DIORA-BMM are able to infer the correct grammars for the structured baseline languages described in the previous section.
In particular, we consider if the induced grammars reflect the correct \emph{word classes} defined by the pre-terminals, and if they capture the simple hierarchy defined on top of these word-classes.

\paragraph{Results} We conclude that CCL-BMM is able to correctly identify all the unique word classes for the examined languages, as well as the simple hierarchy (for some examples of induced grammars, we refer to Appendix~\ref{sec:appendix-struct_lang}).
DIORA-BMM performs well for the smallest languages, but for the most complex grammar ($V=27$, $L=10$) it is only able to find half of the word classes and some of the word class combinations.
We also observe that DIORA-BMM appears to have a bias for binary trees, which results in larger and less interpretable grammars for the longer fully structured languages.
Overall, we conclude that both CCL-BMM and DIORA-BMM should be able to infer interesting grammars for our artificial setup; CCL-BMM appears to be slightly more adequate.

\subsection{Grammar consistency and data size}
As a next step, we study the impact of the induction set sample on the resulting grammars.
We do so by measuring the \emph{consistency} of grammars induced on different sections of the training data as well as grammars induced on differently-sized sections of the training data.
We consider incrementally larger message pools of size $N=\{500,1000,2000,4000,8000\}$ by sampling from the $V27L10$ language with replacement according to the original message frequencies.
From each pool we take the unique messages to induce the grammar.
More details on this procedure and the resulting data sets can be found in Appendix~\ref{sec:appendix-experiment1}.

We express the consistency between two grammars as the F1-score between their parses on the same test data.
We furthermore consider the GDL of the induced grammars, which we compare with a baseline grammar that contains exactly one prediction rule for each message.
If the GDL of the induced grammar is not smaller than the GDL of this baseline grammar, then the grammar was not more efficient than simply enumerating all messages.

The experiments described above provide information about the sensitivity of the grammar induction techniques on the exact section of the training data as well as the size of the training data that is required to obtain a consistent result.
We use the results to find a suitable data set size for the rest of our experiments.

\begin{figure}
  \centering
  \includegraphics[width=0.5\textwidth]{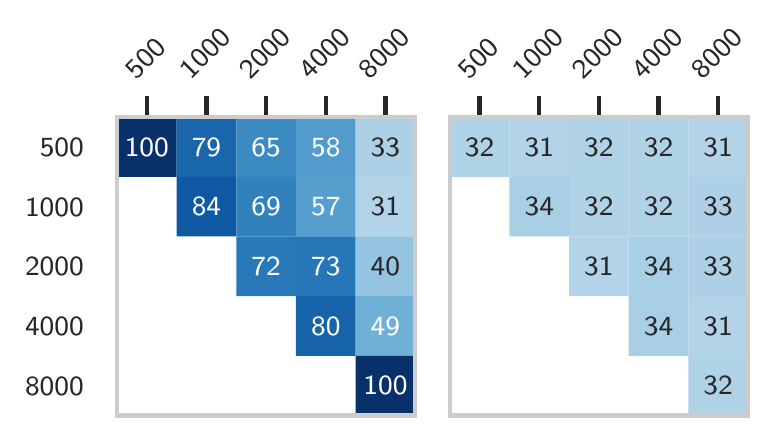}
  \caption{The consistencies for CCL-BMM (left) and DIORA-BMM (right) for language set $V27L10$. The axes show the message pool sizes ($N$) for inducing the compared grammars. %
  }
  \label{fig:consistencies}
\end{figure}

\paragraph{Results} Overall, the experiments show that CCL-BMM has higher consistency scores than DIORA-BMM, but also more variation between different induction set sizes (see Figure \ref{fig:consistencies}).
From the changing consistencies of CCL-BMM with increasing the number of messages, we conclude that differences in data-set size influence its grammar induction considerably. 
We believe that the low consistency scores of DIORA-BMM are due to the strongly stochastic nature of the neural parser.

For both CCL-BMM and DIORA-BMM, the evaluation set coverage increases with the induction set-size, although CCL-BMM reaches a near perfect coverage much faster than DIORA-BMM.
Furthermore, the GDL implies a lower bound for the required induction set size, since the GDL is only smaller than its baseline for $N>2000$ with CCL-BMM, while the crossover point is even larger for DIORA-BMM.
More details on the progressions of the coverage and GDL can be found in the appendix in Figures \ref{fig:exp1-coverage} and \ref{fig:exp1-GDL} respectively.

To conclude, while a small induction set would suffice for CCL, we decide to use all messages of the induction set, because DIORA requires more data for good results, and we see no evidence that this impairs the performance of CCL-BMM.

\section{Analysing emergent languages}
\label{sec:experiment2}

Having verified the applicability of both CCL-BMM and DIORA-BMM, we use them to induce grammars for all languages described in \S\ref{subsec:languages}. %
We analyse the induced grammars and parses, comparing with the structured, shuffled, and random baselines introduced in \S\ref{subsec:baselines}.

\subsection{Grammar aptitude and compressibility}
We first quantitatively evaluate the grammars, considering the description lengths and their evaluation and overgeneration coverage, as described in \S\ref{subsec:evaluation}.
As a general observation, we note that the GDL increases with the vocabulary size.
This is not surprising, as larger vocabularies require a larger number of lexical rules and allow for more combinations of symbols, but indicates that comparisons across different types of languages should be taken with care.

\subsubsection{L3 and L5} 
As a first finding, we see that little to no structure appears to be present in the shorter languages with messages of length 3 and 5: there are no significant differences between the emergent languages and the random and shuffled baseline (full plots can be found in the appendix, Figures~\ref{fig:GDL} and \ref{fig:DDL}).
Some of the grammars for the emergent $L3$ languages and random baselines, however, have a surprisingly low GDL.
Visual inspection of the trees suggests that this is due to the fact that the grammars approach a trivial form, in which there is only one pre-terminal $X$ that expands to every lexical item in the corpus, and one production rule $S \rightarrow X X X$.\footnote{In the case of DIORA-BMM, it is a trivial \emph{binary} tree $S \rightarrow A X$ and $A \rightarrow X X$.}
This result is further confirmed by the \emph{coverages} presented in Table~\ref{tab:coverage_L10}, which illustrates that the grammars for the $L3$ and $L5$ languages can parse not only all sentences in these languages, but also all other possible messages with the same length and vocabulary.

Interestingly, for DIORA-BMM, there are also no significant differences for the \emph{structured} baselines.
We hypothesise that this may stem from DIORA's inductive bias and conclude that for the analysis of shorter languages, CCL-BMM might be more suitable.

\subsubsection{L10}

\begin{figure*}
  \includegraphics[width=\textwidth]{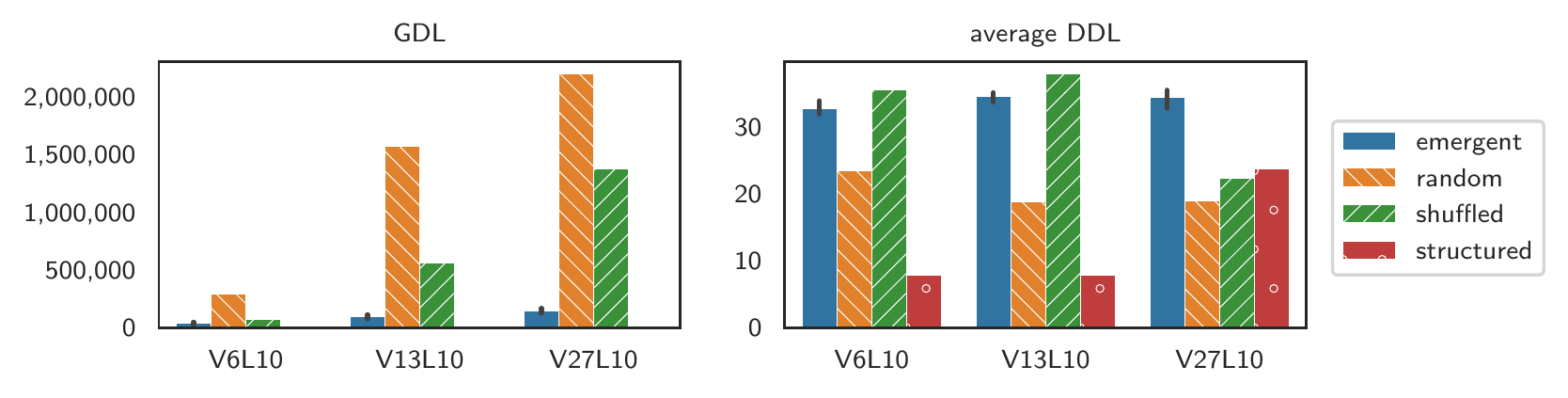}
  \caption{The grammar description lengths (GDL) and average data description lengths (DDL) for the CCL-BMM induced grammars with $L=10$. The languages with $L=\{3,5\}$ and the DIORA-BMM induced grammars are left out and can be found in Figures \ref{fig:GDL} and \ref{fig:DDL}. The GDL of the \emph{structured baseline} is too small to be seen.}
  \label{fig:GDL_DDL_L10}
\end{figure*}

In the $L10$ languages, we find more indication of structure.
As can be seen in Figure \ref{fig:GDL_DDL_L10}, the emergent grammars differ all significantly from all baselines grammars ($p<.05$) and most strongly from the \emph{random baseline} ($p<.001$).
The GDL of the shuffled baseline grammar is in-between the language and random baseline grammar, suggesting that some regularity may be encoded simply in the frequency distribution of the symbols.

The average DDL of the $L10$ languages, however, also differs considerably from the baselines, but in the other direction: both the structured and the completely random baseline are much smaller than the emergent language DDL.
An explanation for this discrepancy is suggested when looking at their \emph{coverages}.
A good grammar has a high coverage on an independent evaluation set with messages from the same language, but a low coverage on a random sample of messages outside of the language (which we measure with \emph{overgeneration coverage}, see \S\ref{subsec:evaluation}).
A perfect example of such a grammar is the CCL-BMM grammar inferred for the structured baseline, which has a coverage of $100\%$ for the evaluation set but approximately $0\%$ outside of it (see Table~\ref{tab:coverage_L10}).
For the $V13L10$ and $V27L10$ languages, we observe a similar pattern.

Coming back to the random languages, we can see that their grammars do not generalise to \emph{any} message outside of their induction set.
This result suggests that for these languages, the induction method resulted in a large grammar that keeps the DDL low at the expense of a larger GDL, by simply overfitting to exactly the induction set.

Concerning the coverage, another interesting finding is that the shuffled baseline often has a higher coverage than the random baseline.
Combined with the generally higher average DDL, this suggests that the induction methods are less inclined to overfit the shuffled baselines.
This might be explained by the regularities present in the shuffled messages through the frequencies of the symbols, as well as their co-occurrences within messages.

\begin{table}
    \setlength\tabcolsep{4pt}
  \centering
  { \small
  \begin{tabular}{ll|cc|cccccc}
\toprule
  & {} & \multicolumn{2}{l|}{evaluation (\%)} & \multicolumn{4}{l}{overgeneration (\%)} \\
L & V &                emerg. & struct. &                 emerg. & rand. & shuf. & struct. \\
    \midrule
    3  & 6  &                100 &   100 &                 100 & 100 & 100 &   0 \\
   & 13 &                100 &   100 &                 100 & 100 & 100 &   0 \\
   & 27 &                100 &   100 &                 100 & 100 & 100 &   0 \\
5  & 6  &                100 &   100 &                 100 & 100 & 100 &   0 \\
   & 13 &                100 &   100 &                 100 & 100 & 100 &   0 \\
   & 27 &                100 &   100 &                 100 & 100 & 100 &   0 \\
10 & 6  &                100 &   100 &                  78$\pm$2 &   0 &  94 &   0 \\
   & 13 &                 \textbf{98$\pm$1} &   100 &                   \textbf{3$\pm$1} &   0 &  13 &   0 \\
    & 27 &                \textbf{96$\pm$1} &   100 &                   \textbf{1$\pm$1} &   0 &   0 &   0 \\
    \bottomrule
  \end{tabular}
  }
  \caption{The average evaluation and overgeneration coverage for the CCL-BMM induced grammars. In bold we emphasise where we recognise a pattern of high evaluation coverage, but low overgeneration coverage. Standard deviations of $<0.5$ for the emergent languages are left out.}
  \label{tab:coverage_L10}
  \end{table}

\subsection{Nature of syntactic structure}
The description lengths and coverage give an indication of whether there is \emph{any} structure present in the languages, we finish with an explorative analysis of the \emph{nature} of this structure.
We focus our analysis on the $V13L10$ and $V27L10$ languages, which we previously found most likely to contain interesting structure.

\subsubsection{Word class structure}
We first examine if there is any structure at the lexical level, in the form of \emph{word classes}. 
We consider the number of terminals per pre-terminal and vice versa.
We will discuss the most important results here, the complete results can be found in the appendix, in Figure \ref{fig:wordclass}.

A first observation is that in all grammars each symbol is unambiguously associated with only one pre-terminal symbol, indicating that there is no ambiguity with respect to the word class it belongs to.
The number of terminals per pre-terminal suggests that our grammar induction algorithms also do not find many word classes: with some notable exceptions, every pre-terminal symbols expand only to a single terminal symbol.
Interestingly, some of these exceptions overlap between CCL-BMM and DIORA-BMM (see Table~\ref{tab:shared_wordclass}), suggesting that they in fact are indicative of some form of lexical structure.

\begin{table}
  \setlength\tabcolsep{4pt}
  \centering
  {\small
    \begin{tabular}{l|ll}
      \toprule
      seed & CCL-BMM & DIORA-BMM \\
      \midrule
      0 & $\{14, \mathbf{16}, \mathbf{24}\}$ & $\{\mathbf{16}, 19, \mathbf{24}\}$ \\
      1 & $\{0, \mathbf{10}\}$      & $\{\mathbf{10}, 22\}$ \\
      2 & none             & $\{0, 18\}$ \\
      \bottomrule
    \end{tabular}
  }
  \caption{An overview of the captured word classes found in language $V27L10$ by CCL-BMM and DIORA-BMM. The overlap between the word-classes found by both setups is indicated in bold.}
  \label{tab:shared_wordclass}
\end{table}

\subsubsection{Higher level structure}
\begin{table}
  \setlength\tabcolsep{4pt}
  \centering
  {\small
\begin{tabular}{lc|cccc}
\toprule
 L & V &             emergent &   random &   shuffled & structured \\
  \midrule
  10 & 6  &          34.7 $\pm$0.9 &  36.0 &  36.0 &    2.0* \\
  & 13 &             78.7 $\pm$6.0 &  169* &  137* &    2.0* \\
  & 27 &             192\hspace{2mm} $\pm$65 &  441* &  262  &    2.0 \\
\bottomrule
\end{tabular}
}

\caption{The number of unique \emph{pre-terminal groups} in the CCL-BMM induced grammars for $L=10$. A pre-terminal group constitutes the right-hand side of a production rule leading only to pre-terminals or symbols.
An asterisk (*) indicates a significant difference with the baseline value (p < .05).
}
\label{tab:preterminal_groups_ccl}
\end{table}

We next check if the trees contain structure one level above the pre-terminals, by computing if pre-terminals can be grouped based on the non-terminal that generates them (e.g.\ if there is a rule \emph{K $\rightarrow$ A B} we say that \emph{K} generates the group \emph{A B}). 
Specifically, we count the unique number of \emph{pre-terminal groups}, defined by each right-hand side consisting solely of pre-terminals and symbols.
If there is an underlying linguistic structure that prescribes which pre-terminals belong together (and in which order), it is expected that fewer groups are required to explain the messages than if no such hierarchy were present.
Indeed, the number of \emph{pre-terminal groups} (see Table \ref{tab:preterminal_groups_ccl}) shows this pattern, as we discover a significantly smaller number of groups than the random baseline.
These results thus further confirm the presence of structure in the $V13L10$ and $V27L10$ languages.

As a tentative explanation, we would like to suggest that perhaps the symbols in the emergent languages are more akin to characters than to words.
In that case, the pre-terminal groups would represent the words, and the generating non-terminals the word-classes.
For both CCL-BMM and DIORA-BMM, the average number of pre-terminal groups generated by these non-terminals is $2.4 \pm <0.01$ for the emergent languages, while it is $1.0$ for the shuffled and random baselines. 
This suggests that the pre-terminal groups share in syntactic function. 
Such observations could form a fruitful basis for further semantic analysis of the languages.

\begin{figure*}[!h]
  \begin{subfigure}{0.5\textwidth}
    \centering
    \includegraphics[width=0.98\textwidth]{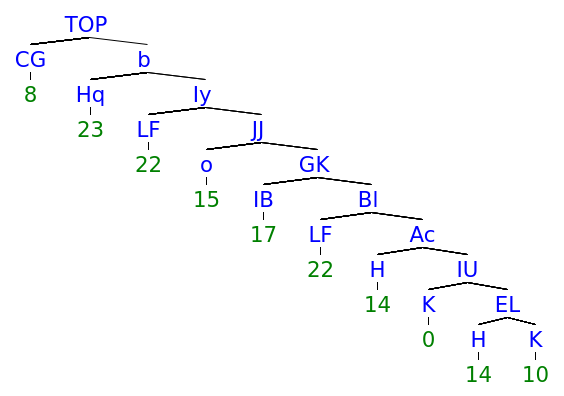}
  \end{subfigure}
  \begin{subfigure}{0.5\textwidth}
    \centering
    \includegraphics[width=0.98\textwidth]{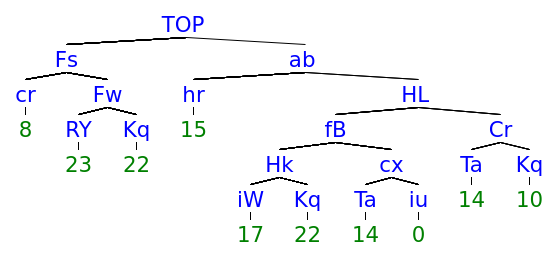}
  \end{subfigure}
  \caption{Example parse trees from the same $V27L10$ evaluation set by a CCL-BMM (left) and DIORA-BMM (right) induced grammar. It should be noted that this is one of the few exceptions for $L=10$ where some symbols share a pre-terminal.}
  \label{fig:example_parsetree}
\end{figure*}

\subsubsection{Recursion}
Lastly, we would like to note the lack of recursive production rules in nearly all induced grammars. 
While this is not surprising given both the previous results as well as the simplicity of the meaning space, it does suggest that perhaps more interesting input scenarios are required for referential games.

\subsection{CCL vs DIORA}
We ran all our experiments with both CCL-BMM and DIORA-BMM.
There were similarities, but also some notable differences.
Based on the GDL, CCL-BMM seems more suitable to analyse shorter languages, but earlier tests with reconstructing the structured baseline grammars (see \S\ref{subsec:struct_grammars}) suggest that DIORA-BMM also performs worse on languages with a \emph{larger} message length and vocabulary size; leading us to believe that CCL-BMM is more appropriate for our setup.

Another difference concerns the distribution of the tree depths, which reflects mostly skewed and binary trees for CCL-BMM for $L=10$, but more evenly distributed depths for DIORA-BMM (for a plot of the depth distributions, we refer to \ref{fig:tree_distributions}).
An example of this difference is shown in Figure \ref{fig:example_parsetree}.
A possible explanation is that CCL-BMM is more biased towards fully right-branching syntax trees, since these are a good baseline for natural language.
Alternatively, these trees might actually reflect the emergent languages best, perhaps because of the left-to-right nature of the agents' LSTMs.
Additional work is required to establish which type of trees better reflect the true structure of the emergent languages.

\section{Conclusion}

While studying language and communication through \emph{referential games} with \emph{artificial agents} has recently regained popularity, there is still a very limited amount of tools available to facilitate the analysis of the resulting emergent languages.
As a consequence, we still have very little understanding of what kind of information these languages encode. %
In this paper, for the first time, we focus on \emph{syntactic} analysis of emergent languages.

We test two different unsupervised grammar induction (UGI) algorithms that have been successful for natural language: a pre-neural statistical one, CCL, and a neural one, DIORA.
We use them to infer grammars for a variety of languages emerging from a simple referential game and then label those trees with BMM, considering in particular the effect of the message length and vocabulary size on the extent to which structure emerges.

We first confirm that the techniques are capable of inferring interesting grammars for our artificial setup and demonstrate that CCL appears to be a more suitable constituency parser than DIORA.
We then find that the shorter languages, with messages up to 5 symbols, do not contain any interesting structure, while languages with longer messages appear to be substantially more structured than the two random baselines we compare them with.
Interestingly, our analysis shows that even these languages do not appear to have a notion of \emph{word classes}, suggesting that their symbols may in fact be more akin to \emph{letters} than to words.
In light of these results, it would be interesting to explore the use of unsupervised tokenisers that work well for languages without spaces \citep[e.g.\ SentencePiece][]{kudo2018sentencepiece} prior to our approach and to try other word embedding models for DIORA, such as the \emph{character-}based ELMo embeddings\footnote{DIORA already supports ELMo vectors besides GloVe.} \cite{peters2018deep} or the more recent BERT \cite{devlin2018bert}.

Our results also suggest that more sophisticated game scenarios may be required to obtain more interesting structure.
UGI could provide an integral part in analysing the languages emerging in such games, especially since it -- contrary to most techniques previously used for the analysis of emergent languages -- does not require a description of the hypothesised semantic content of the messages.
Examples of more sophisticated game scenarios are bidirectional conversations where multi-symbol messages are challenging to analyse \cite{kottur2017natural,bouchacourt2019miss} or games with image sequences as input \cite{santamaria2019towards}.

We argue that while the extent to which syntax develops in different types of referential games is an interesting question in its own right, a better understanding of the syntactic structure of emergent languages could also provide pivotal in better understanding their \emph{semantics}, especially if this is considered from a compositional point of view.
To facilitate such analysis, we bundled our tests in a comprehensive and easily usable evaluation framework.\footnote{\url{https://github.com/i-machine-think/emergent_grammar_induction}}
We hope to have inspired other researchers to apply syntactic analysis techniques and encourage them to use our code to evaluate new emergent languages trained in other scenarios.

\section*{Acknowledgments}

The idea to use grammar induction techniques to analyse emergent languages arose during an internship at FAIR Paris. We thank Marco Baroni and Diane Bouchacourt for the fruitful discussions about this topic that took place during that time.
DH is funded by the Netherlands Organization for Scientific Research (NWO), through a Gravitation Grant 024.001.006 to the Language in Interaction Consortium.
EB the European Union's Horizon 2020 research and innovation program under the Marie Sklodowska-Curie grant agreement No 790369 (MAGIC).

\bibliography{references}
\bibliographystyle{acl_natbib}

\clearpage

\appendix
\renewcommand\thefigure{\thesection.\arabic{figure}} 
\renewcommand\thetable{\thesection.\arabic{table}}
\setcounter{table}{0}
\setcounter{figure}{0}

\section{Definition of the referential game}
\label{sec:appendix-ref_game}

The languages emerge from two agents playing a referential game with a setup similar to \citet{havrylov2017emergence}.
In each round of the game, the sender samples a message $m$ describing the target image $t$ to the receiver.
$m$ consists of up to $L$ symbols sampled from a vocabulary with size $V$.\footnote{Technically, the vocabulary also contains a \emph{stop character} and the sender is allowed to generate messages shorter than $L$. However, typically the messages have a length of $L$. For the analyses in this paper we have removed all stop characters in a pre-processing step and we do not count it as part of $L$ and $V$.}
The receiver has to identify the described image from a set with $t$ and three other distracting images in random order.
The images are created by generating a shape with a certain colour and size, on a logical grid.
In the game, two images are the same if they have the same colour, shape, and size, even when differently positioned.
Table \ref{tab:agent_design} provides an overview of the agents' architectures used in this game.

\begin{table}[!h]
  {\footnotesize
  \begin{tabular}{l|lr}
    \toprule
    LSTM     & Embedding size                 & $256$ \\
    {}       & Hidden layer size              & $512$ \\
    \midrule
    CNN      & \# of convolutional layers     & $5$ \\
    {}       & \# of filters                  & $20$ \\
    {}       & Kernel size                    & $3$ \\
    {}       & Stride                         & $2$ \\
    {}       & No padding                     & {} \\
    {}       & Activation function            & ReLU \\
    \bottomrule
  \end{tabular}
  }
  \caption{Parameters for the sender and receiver architecture. The convolutional layers are followed by batch normalisation.}
  \label{tab:agent_design}
\end{table}

\setcounter{table}{0}
\setcounter{figure}{0}
\section{Fully structured languages}
\label{sec:appendix-struct_lang}

For all the configurations of $L$ and $V$ of our emergent languages (see \S\ref{subsec:languages}), we create a simple grammar containing word classes, each with a disjoint set of symbols.
Furthermore, two pre-terminals form a group that can be placed either at the beginning or the end of the message or both, while the other pre-terminals occupy the remaining spots in fixed order.
The smaller grammars repeat word classes to ensure enough messages for the induction and evaluation.

\begin{table}[!h]
  \centering
  {  \footnotesize
  \begin{tabular}{ll|rrr}
\toprule
L  & V  & total                    & induction     & evaluation            \\
    \midrule
3  & 6  &                16 & 12 & 4 \\
   & 13 &                160 & 128 & 32 \\
   & 27 &                1458 & 1166 & 292 \\
5  & 6  &                24 & 19 & 5 \\
   & 13 &                378 & 302 & 76 \\
   & 27 &                15480 & 2000 & 500 \\
10 & 6  &                24 & 19 & 5 \\
   & 13 &                32 & 25 & 7 \\
   & 27 &                52488 & 2000 & 500 \\
    \bottomrule
  \end{tabular}
  }
  \caption{An overview of the total number of possible messages that can be generated for each $L$ and $V$ configuration, as well as the sizes of the induction and evaluation sets.
    The size of the induction set is capped at $2000$ to keep the grammar induction computationally feasible. When evaluating the grammars a maximum number of $500$ messages of either set is used.}
  \label{tab:dataset-4.1}
\end{table}

All the possible messages are randomly divided over a induction and evaluation set ($80\%$ and $20\%$ respectively).
Table \ref{tab:dataset-4.1} provides more details on the data sets used for each language configuration.

\subsection{Example grammars}
In the following examples, \texttt{TOP} denotes the start symbol, \texttt{NP} the pre-terminal group, and the numbers the terminals that represent the symbols in the generated messages.

The structured baseline grammar for $V=13$ and $L=5$ is represented as:
{  \footnotesize
\begin{verbatim}
    TOP -> NP AP
    TOP -> AP NP
    TOP -> NP VP NP
    NP  -> A B
    AP  -> E C D
    VP  -> E
    A   -> 0 | 1 | 2
    B   -> 3 | 4 | 5
    C   -> 6 | 7 | 8
    D   -> 9 | 10
    E   -> 11 | 12
\end{verbatim}
  }

The resulting CCL-BMM induced grammar is:
{  \footnotesize
\begin{verbatim}
    TOP -> H E A
    TOP -> A G
    G   -> H A | H E
    E   -> B F
    A   -> C D
    C   -> 0 | 1 | 2
    D   -> 3 | 4 | 5
    B   -> 6 | 7 | 8
    F   -> 9 | 10
    H   -> 11 | 12
\end{verbatim}
  }

and DIORA-BMM finds:
{  \footnotesize
\begin{verbatim}
    TOP -> K A
    TOP -> L D
    TOP -> B J
    TOP -> N H
    TOP -> J B
    E -> C K
    A -> O D | F H | D J
    G -> B C | K F
    O -> F K | D E
    F -> D C
    H -> M I
    L -> J K | B E | G K | K O
    B -> K D
    J -> E D | C B | C H
    N -> K F
    K -> 0 | 1 | 2
    D -> 3 | 4 | 5
    M -> 6 | 7 | 8
    I -> 9 | 10
    C -> 11 | 12
\end{verbatim}
  }

\setcounter{table}{0}
\setcounter{figure}{0}
\section{Consistency and suitable data set size}
\label{sec:appendix-experiment1}

The number of messages in the induction set might influence the properties of the grammars induced from it.
To investigate these effects, we perform induction experiments on different sub-samples of the language $V27L10$. %
We compare the induced grammars on their consistency and study the progression of the evaluation coverage and GDL.

The consistency of a setup is computed on different samples of a data set to study the effect of the data set size as well as to show how dependent the algorithm is on the exact selection of induction messages.
We create incrementally larger pools by sampling a fixed number of randomly selected messages from the data-set, resulting in pool sizes $N=\{500,1000,2000,4000,8000\}$.
The messages are sampled with replacement according to the frequency in the original language.
From these pools we then only consider the unique messages.
The procedure is repeated three times for each $N$ to obtain an average consistency.

\begin{figure}
  \centering
  \includegraphics[width=.45\textwidth]{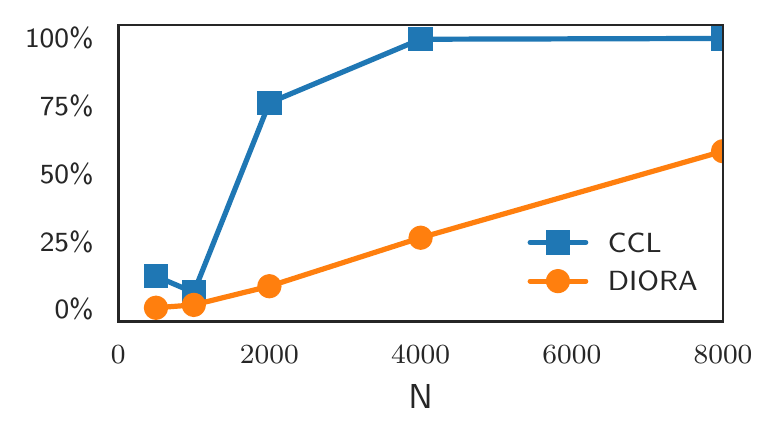}
  \caption{Average evaluation coverage of the CCL-BMM and DIORA-BMM induced grammars ($V27L10$) against the induction pool size $N$.}
  \label{fig:exp1-coverage}
\end{figure}

\begin{figure}[H]
    \centering\includegraphics[width=.45\textwidth]{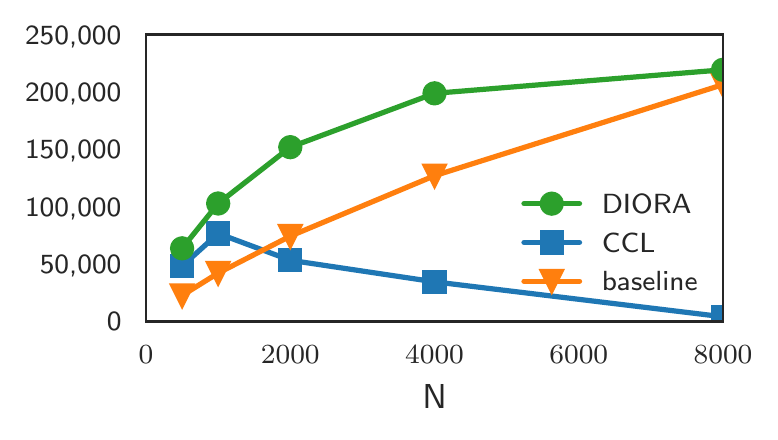}
    \caption{Progression of the average GDL of the induced grammars ($V27L10$) compared to the baseline grammar of one production rule for each message.}
    \label{fig:exp1-GDL}
  \end{figure}

Subsequently, we study the average evaluation coverage and GDL for these grammars.
The resulting progression of the evaluation coverage is shown in Figure \ref{fig:exp1-coverage}.
The coverage is evaluated with respect to the disjoint set consisting of $10\%$ of the language's messages.
We study the GDL against the number of messages compared to the baseline grammar of one production rule for each message in the induction set in Figure \ref{fig:exp1-GDL}.

\setcounter{table}{0}
\setcounter{figure}{0}
\section{Analysing emergent languages}
\label{sec:appendix-experiment2}

Here we present a complete overview of the results from analysing the languages in \S\ref{sec:experiment2}. To aid in interpreting the different metrics, we compare these with several baselines. To test for significance, we report the \emph{p}-values from a \emph{one-sample t-test}, where the baseline value is assumed to be the population mean.

\subsection{Baselines}
\label{subsec:appendix-baselines}

The \emph{shuffled baselines} are constructed by randomly shuffling the messages of the \emph{induction set} for a randomly selected seed, such that they are unique in the shuffled set. We create the \emph{random baselines} by randomly sampling the same number of unique messages as the \emph{induction set}, also for one seed. See Table \ref{tab:baselines} for the number of messages used for each baseline per language.

\begin{table}[]
  \centering
  {  \footnotesize
  \begin{tabular}{ll|rr}
    \toprule
    L  & V  & shuffled & random \\
    \midrule
3	&	6	&	147	&	150	\\
	&	13	&	358	&	396	\\
	&	27	&	512	&	554	\\
5	&	6	&	913	&	829	\\
	&	13	&	1819	&	1590	\\
	&	27	&	1962	&	1817	\\
10	&	6	&	4266	&	4525	\\
	&	13	&	8248	&	8294	\\
	&	27	&	9112	&	8986	\\
    \bottomrule
  \end{tabular}
}
\caption{Number of messages per language for the shuffled and random baseline.}
  \label{tab:baselines}
\end{table}

\subsection{Description lengths}
\begin{figure*}%
    \centering
      \includegraphics[width=\textwidth]{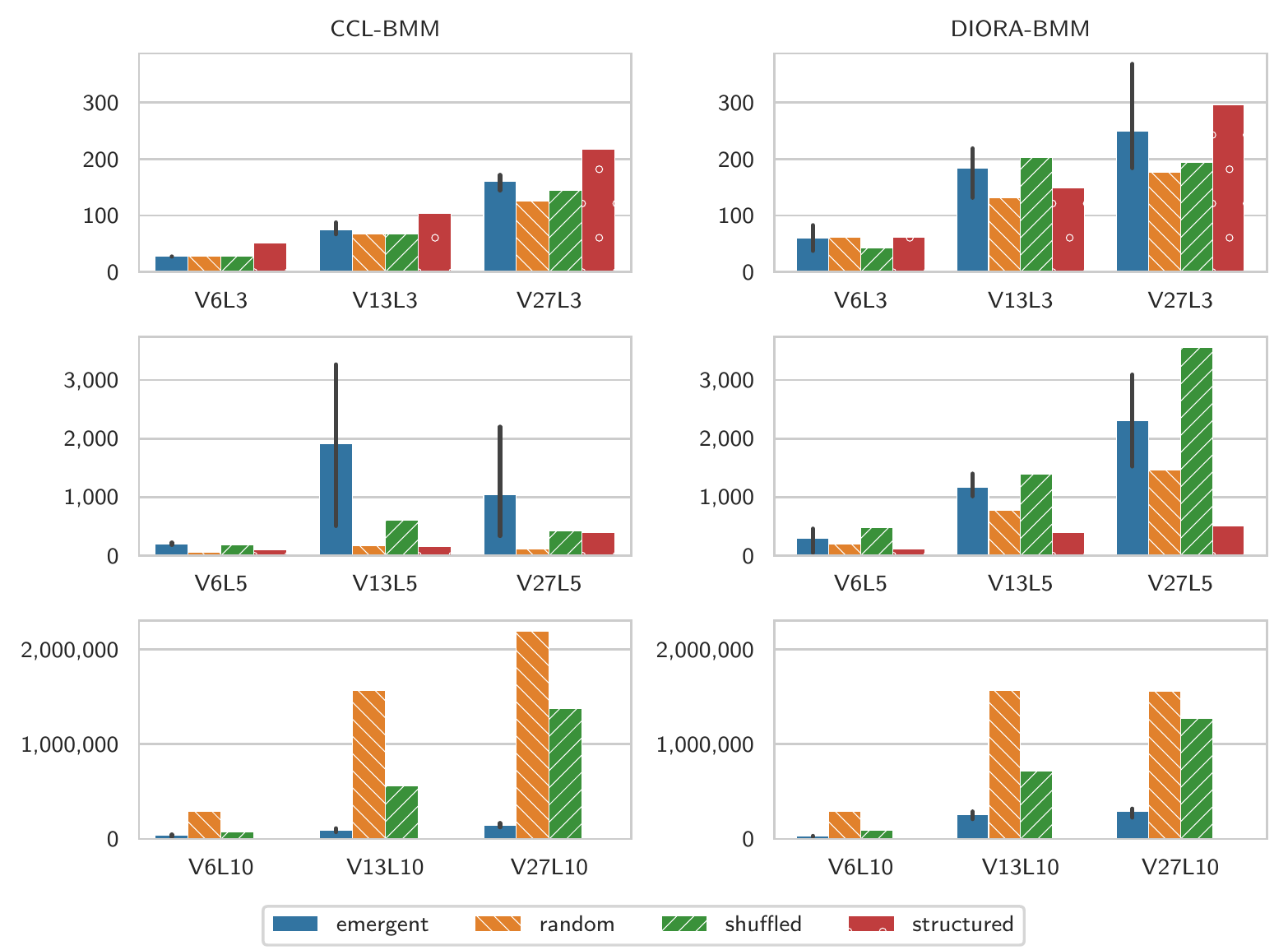}
      \caption{Overview of the grammar description lengths (GDL) for the induced grammars. Note that for $L=10$ the structured baseline GDL is too small to be visible in the chart.}
      \label{fig:GDL}
    \end{figure*}

\begin{figure*}%
      \includegraphics[width=\textwidth]{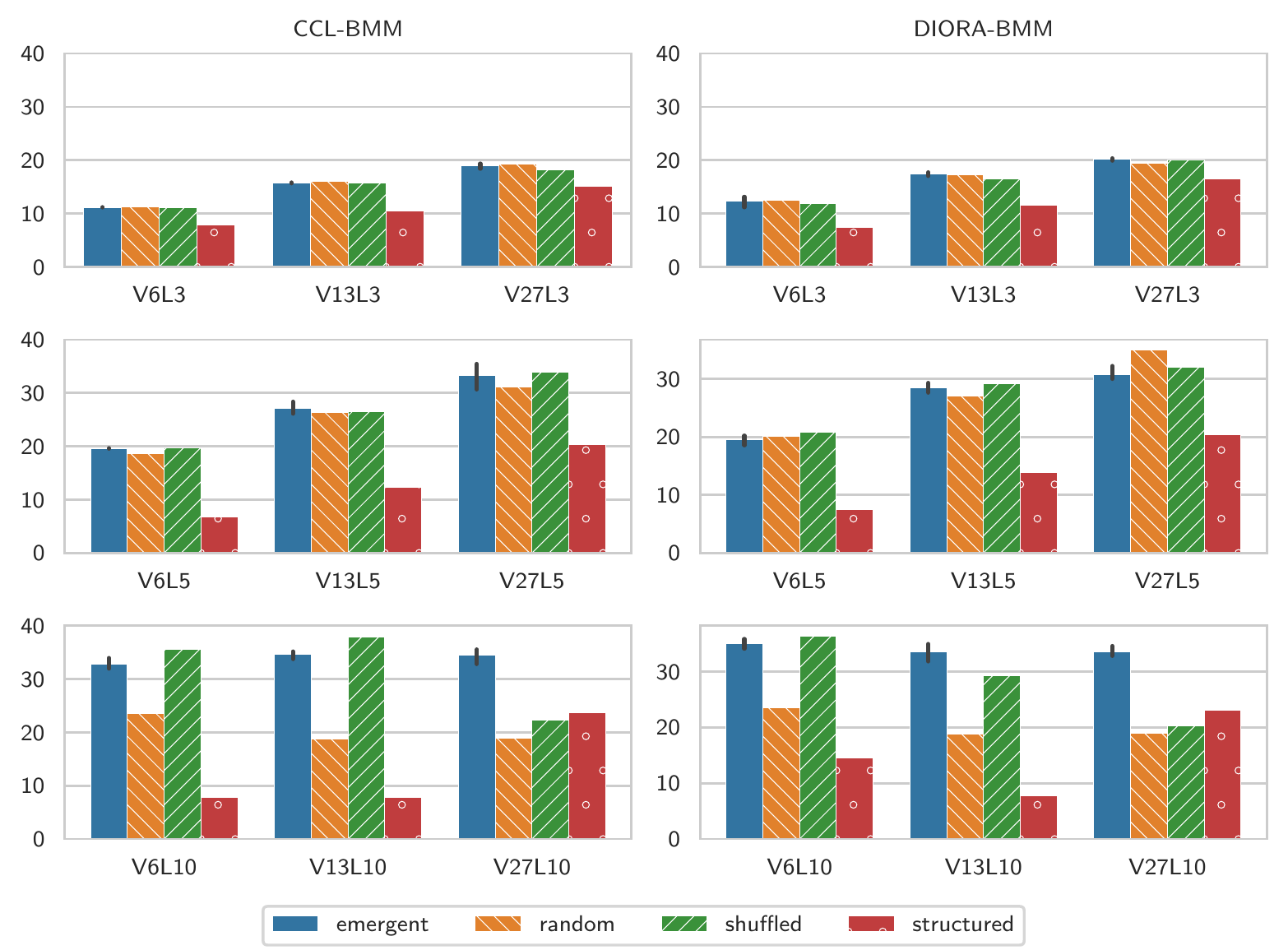}
      \caption{Overview of the data description lengths (DDL) for the induced grammars.}
      \label{fig:DDL}
    \end{figure*}

Tables \ref{tab:description_lengths}, \ref{tab:eval_DDL}, and \ref{tab:ratio_DDL/GDL} give an overview of the description lengths for the induction sets, the evaluation sets, and their ratios, respectively. The description lengths are also visualised in Figures \ref{fig:GDL} and \ref{fig:DDL}.

\subsection{Coverage}
We show the evaluation and overgeneration coverage in Table \ref{tab:coverages}. %

\subsection{Nature of syntactic structure}
\begin{figure*}%
  \centering
  \begin{subfigure}[b]{\textwidth}
    \includegraphics[width=\textwidth]{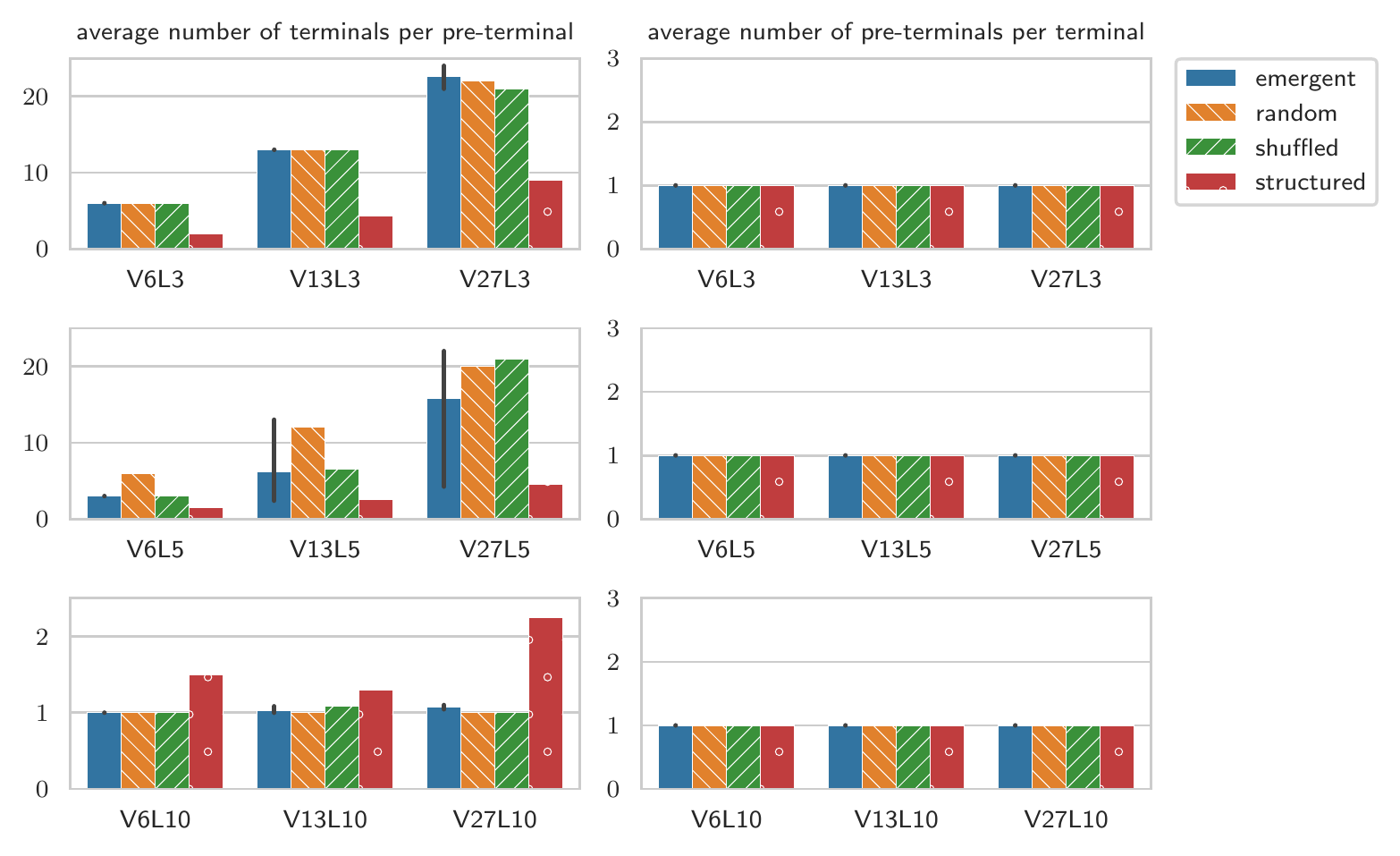}
    \caption{CCL-BMM}
  \end{subfigure}
  \begin{subfigure}[b]{\textwidth}
    \includegraphics[width=\textwidth]{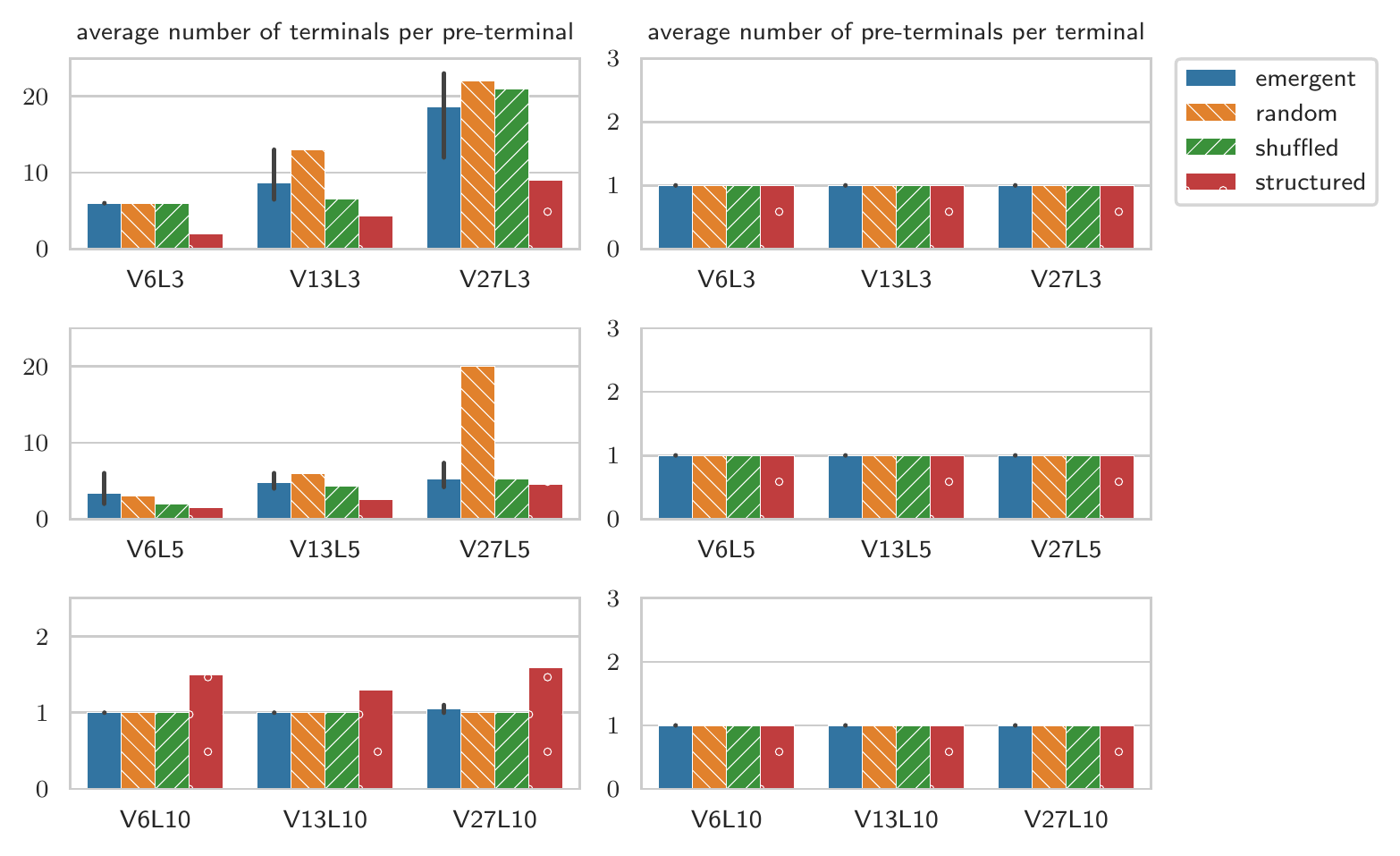}
    \caption{DIORA-BMM}
  \end{subfigure}
  \caption{An overview of the average number of \emph{terminals per pre-terminal} and the average number of \emph{pre-terminals per terminal}.}
  \label{fig:wordclass}
\end{figure*}

Table \ref{tab:number_preterminals} gives an overview of the total number of unique pre-terminals and terminals in the induced grammars. We show the average number of pre-terminals per terminal in Table \ref{tab:wordclass} and Figure \ref{fig:wordclass}. The average number of pre-terminals per terminal is one for every language and baseline, and is therefore omitted. %
The number of pre-terminal groups and the number of non-terminals generating these groups are presented in Table \ref{tab:preterminal_groups}.

\subsection{Parse tree distributions}
\begin{figure*}%
  \begin{subfigure}{0.5\textwidth}
    \includegraphics[width=\textwidth]{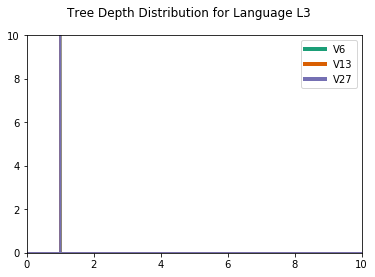}
\end{subfigure}
\begin{subfigure}{0.5\textwidth}
  \includegraphics[width=\textwidth]{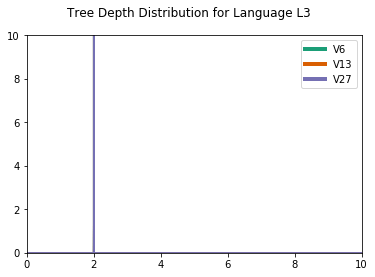}
\end{subfigure}
\begin{subfigure}{0.5\textwidth}
  \includegraphics[width=\textwidth]{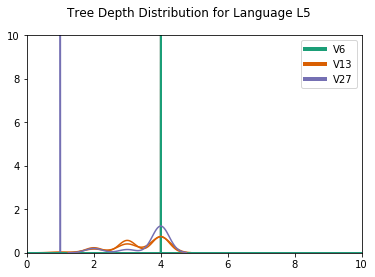}
\end{subfigure}
\begin{subfigure}{0.5\textwidth}
  \includegraphics[width=\textwidth]{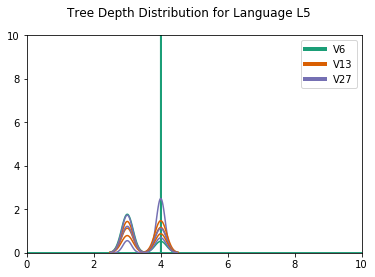}
\end{subfigure}
\begin{subfigure}{0.5\textwidth}
  \includegraphics[width=\textwidth]{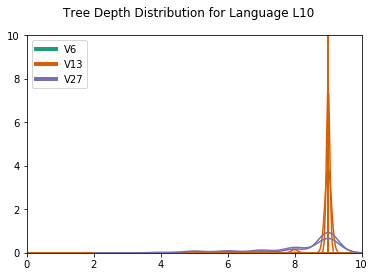}
  \caption{CCL-BMM}
\end{subfigure}
\begin{subfigure}{0.5\textwidth}
  \includegraphics[width=\textwidth]{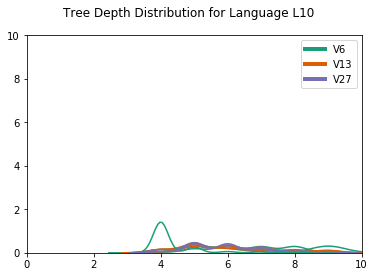}
  \caption{DIORA-BMM}
\end{subfigure}
\caption{Visualisations of the \emph{parse tree depth} distributions for the most probable parses of the evaluation messages for all emergent languages.}
\label{fig:tree_distributions}
\end{figure*}

In Figure \ref{fig:tree_distributions} we show the parse tree distributions.
For the CCL induced $L3$ grammars, we see all depths are 1, while for DIORA all depths are 2. A parse depth of 1 indicates a flat grammar, without hierarchical structure. %
The depth of 2 reflects the bias of DIORA towards binary trees.

The $L5$, and especially $L10$, grammars show deeper trees, often the maximum tree depth, which would mean binary skewed trees. DIORA shows more variation in the tree depth distributions. %

\begin{table*}%
{  \footnotesize
  \begin{subtable}[h]{\textwidth}
\centering
\begin{tabular}{ll|llll|llll}
\toprule
   & {} & \multicolumn{4}{l|}{GDL} & \multicolumn{4}{l}{average DDL} \\
L   & V  &              emergent &          random &          shuffled & structured &       emergent &     random &    shuffled &   structured \\
\midrule
3  & 6  &             28$\pm$0.0 &            28 &            28 &  52* &  11.2$\pm$0.0 &     11.2 &    11.2 &   7.8* \\
   & 13 &            74$\pm$10 &            67 &            67 &   1.0E02* &  15.7$\pm$0.1 &    16.0* &    15.7 &  10.5* \\
   & 27 &           1.6E02$\pm$12 &           1.3E02 &           1.5E02 &   2.1E02* &  18.9$\pm$0.4 &     19.2 &    18.2 &    15.2* \\
5  & 6  &           2.0E02$\pm$18 &           62* &           1.8E02 &    97* &  19.6$\pm$0.0 &  18.6* &    19.7 &   6.8* \\
   & 13 &    1.9E03$\pm$1.1E03 &           1.7E02 &           6.1E02 &    1.6E02 &  27.1$\pm$0.9 &     26.4 &    26.6 &   12.3* \\
   & 27 &    1.0E03$\pm$8.3E02 &           1.1E02 &           4.3E02 &    4.0E02 &  33.3$\pm$2.0 &     31.2 &    34.0 &    20.3* \\
10 & 6  &   3.6E04$\pm$9.4E03 &    2.9E05* &       7.2E04* &   1.3E02* &  32.7$\pm$0.9 &   23.5* &   35.6* &   7.8* \\
   & 13 &   9.3E04$\pm$1.6E04 &  1.6E06* &    5.6E05* &   2.9E02* &  34.6$\pm$0.6 &  18.8* &   38.0* &   7.8* \\
   & 27 &  1.4E05$\pm$1.8E04 &  2.2E06* &  1.4E06* &   9.8E02* &  34.5$\pm$1.2 &   18.9* &  22.3* &    23.8* \\
\bottomrule
\end{tabular}

\caption{CCL-BMM}
\end{subtable}
\hfill
\begin{subtable}[h]{\textwidth}
  \centering
  \begin{tabular}{ll|llll|llll}
\toprule
   & {} & \multicolumn{4}{l|}{GDL} & \multicolumn{4}{l}{average DDL} \\
L   & V  &              emergent &          random &          shuffled &  structured &       emergent &    random &    shuffled &   structured \\
\midrule
3  & 6  &            61$\pm$18 &            62 &            42 &      62 &  12.3$\pm$0.8 &    12.5 &    11.9 &     7.4* \\
   & 13 &           1.9E02$\pm$38 &           1.3E02 &           2.0E02 &     1.5E02 &  17.4$\pm$0.3 &    17.3 &    16.6 &   11.6* \\
   & 27 &           2.5E02$\pm$85 &           1.8E02 &           1.9E02 &     3.0E02 &  20.1$\pm$0.2 &   19.3* &    20.0 &   16.5* \\
5  & 6  &      2.9E02$\pm$1.8E02 &           1.9E02 &           4.8E02 &     1.2E02 &  29.5$\pm$0.7 &    20.1 &    20.9 &    7.4* \\
   & 13 &    1.2E03$\pm$1.7E02 &           7.7E02 &         1.4E03 &    4.0E02* &  28.5$\pm$0.7 &    27.1 &    29.2 &   13.8* \\
   & 27 &    2.3E03$\pm$6.4E02 &         1.4E03 &         3.6E03 &     5.0E02 &  30.8$\pm$1.0 &   35.0* &    32.0 &   20.4* \\
10 & 6  &   2.9E04$\pm$3.1E03 &    2.9E05* &      9.1E04* &    1.3E02* &  35.0$\pm$0.7 &  23.5* &    36.4 &  14.5* \\
   & 13 &  2.6E05$\pm$3.3E04 &  1.6E06* &     7.2E05* &    3.9E02* &  33.6$\pm$1.3 &  18.8* &   29.3* &    7.8* \\
   & 27 &  2.9E05$\pm$4.3E04 &  1.6E06* &  1.3E06* &  2.9E03* &  33.5$\pm$0.8 &  18.9* &  20.2* &   23.0* \\
\bottomrule
\end{tabular}
\caption{DIORA-BMM}
\end{subtable}
}
\caption{Description Lengths (GDL and average DDL) for the induced grammars and their baselines. We indicate significant differences with the baseline value at $p<.05$ with an asterisk (*).}
\label{tab:description_lengths}
\end{table*}

\begin{table*}%
{  \footnotesize
  \begin{subtable}[t]{0.45\textwidth}
\centering
\begin{tabular}{ll|ll}
\toprule
L   & V  &       emergent &   structured \\
\midrule
3  & 6  &  11.2$\pm$0.0 &   8.2* \\
   & 13 &  15.9$\pm$0.0 &  10.8* \\
   & 27 &  19.1$\pm$0.5 &    15.2* \\
5  & 6  &  19.6$\pm$0.1 &   7.7* \\
   & 13 &  27.0$\pm$0.9 &   12.4* \\
   & 27 &  33.6$\pm$1.8 &    20.3* \\
10 & 6  &  32.9$\pm$0.9 &   8.0* \\
   & 13 &  35.3$\pm$0.5 &   8.4* \\
   & 27 &  35.6$\pm$1.3 &    23.8* \\
\bottomrule
\end{tabular}

\caption{CCL-BMM}
\end{subtable}
\hfill
\begin{subtable}[t]{0.45\textwidth}
\centering
\begin{tabular}{ll|ll}
\toprule
L   & V  &       emergent &   structured \\
\midrule
3  & 6  &  12.4$\pm$0.8 &     7.6* \\
   & 13 &  17.6$\pm$0.3 &   11.9* \\
   & 27 &  20.3$\pm$0.2 &   16.5* \\
5  & 6  &  19.5$\pm$0.7 &    7.7* \\
   & 13 &  28.4$\pm$0.6 &  13.7* \\
   & 27 &  30.8$\pm$1.2 &    20.5* \\
10 & 6  &  35.0$\pm$0.6 &  13.7* \\
   & 13 &  35.7$\pm$1.4 &    8.5* \\
   & 27 &  35.9$\pm$1.4 &    23.1* \\
\bottomrule
\end{tabular}

\caption{DIORA-BMM}
\end{subtable}
}
\caption{Average data description lengths on the evaluation set (average evaluation DDL) for the grammars induced on the languages and their structured baselines.  We indicate significant differences with the baseline value at $p<.05$ with an asterisk (*).}
\label{tab:eval_DDL}
\end{table*}

\begin{table*}%
  \begin{subtable}[t]{\textwidth}
    \centering
{\footnotesize
    \begin{tabular}{ll|rrrr|rr}
\toprule
   &  & \multicolumn{4}{l|}{DDL:GDL} & \multicolumn{2}{l}{evaluation DDL:GDL} \\
L   & V  &    emergent &    random &    shuffled &  structured &         emergent & structured \\
\midrule
3  & 6  &   60.87 &   60.99 &   59.66 &    1.83 &         7.07 &   0.64 \\
   & 13 &   85.67 &   94.42 &   83.70 &   12.99 &         9.69 &   3.32 \\
   & 27 &   65.60 &   84.57 &   64.42 &   81.56 &         7.41 &  20.51 \\
5  & 6  &   83.09 &  249.42 &   98.04 &    1.33 &         9.29 &   0.40 \\
   & 13 &   44.72 &  249.45 &   79.70 &   23.92 &         4.98 &   6.05 \\
   & 27 &  115.86 &  494.86 &  156.57 &  102.85 &        12.99 &  25.71 \\
10 & 6  &    4.39 &    0.37 &    2.10 &    1.12 &         0.49 &   0.30 \\
   & 13 &    3.15 &    0.10 &    0.56 &    0.68 &         0.36 &   0.20 \\
   & 27 &    2.18 &    0.08 &    0.15 &   48.43 &         0.25 &  12.12 \\
\bottomrule
    \end{tabular}}
  \caption{CCL-BMM}
  \end{subtable}
  \begin{subtable}[t]{\textwidth}
    \centering
  {\footnotesize
\begin{tabular}{ll|rrrr|rr}
\toprule
   & {} & \multicolumn{4}{l|}{DDL:GDL} & \multicolumn{2}{l}{evaluation DDL:GDL} \\
L   & V &    emergent &   random &   shuffled & structured &         emergent & structured \\
\midrule
3  & 6  &   33.50 &  30.41 &  41.29 &   1.43 &         3.91 &   0.49 \\
   & 13 &   38.57 &  51.90 &  29.18 &   9.94 &         4.37 &   2.55 \\
   & 27 &   49.52 &  60.86 &  52.72 &  64.78 &         5.57 &  16.29 \\
5  & 6  &  127.10 &  86.48 &  39.66 &   1.21 &        14.22 &   0.33 \\
   & 13 &   39.15 &  56.20 &  38.28 &  10.52 &         4.35 &   2.63 \\
   & 27 &   25.38 &  43.59 &  17.63 &  81.71 &         2.83 &  20.49 \\
10 & 6  &    5.54 &   0.37 &   1.70 &   2.09 &         0.62 &   0.52 \\
   & 13 &    1.10 &   0.10 &   0.34 &   0.50 &         0.13 &   0.15 \\
   & 27 &    1.06 &   0.11 &   0.15 &  15.79 &         0.13 &   3.96 \\
\bottomrule
\end{tabular}
}
\caption{DIORA-BMM}
\end{subtable}
\caption{An overview of the ratios of DDL:GDL and \emph{evaluation DDL}:GDL for all the languages and their baselines.}
\label{tab:ratio_DDL/GDL}
\end{table*}

\begin{table*}%
  {\footnotesize
    \begin{subtable}[h]{\textwidth}
      \centering
  \begin{tabular}{ll|rr|rrrrrr}
\toprule
  & {} & \multicolumn{2}{l|}{evaluation (\%)} & \multicolumn{4}{l}{overgeneration (\%)} \\
L    & V  &                emergent & structured &                 emergent & random & shuffled & structured \\
    \midrule
    3  & 6  &                100 &   100 &                 100 & 100 & 100 &   0 \\
   & 13 &                100 &   100 &                 100 & 100 & 100 &   0 \\
   & 27 &                100 &   100 &                 100 & 100 & 100 &   0 \\
5  & 6  &                100 &   100 &                 100 & 100 & 100 &   0 \\
   & 13 &                100 &   100 &                 100 & 100 & 100 &   0 \\
   & 27 &                100 &   100 &                 100 & 100 & 100 &   0 \\
10 & 6  &                100 &   100 &                  78$\pm$2 &   0 &  94 &   0 \\
   & 13 &                 98$\pm$1 &   100 &                   3$\pm$1 &   0 &  13 &   0 \\
    & 27 &                96$\pm$1 &   100 &                   1$\pm$1 &   0 &   0 &   0 \\
    \bottomrule
  \end{tabular}
  \caption{CCL-BMM}
\end{subtable}
\hfill
\begin{subtable}[h]{\textwidth}
  \centering
\begin{tabular}{ll|rr|rrrr}
\toprule
  & {} & \multicolumn{2}{l|}{evaluation  (\%)} & \multicolumn{4}{l}{overgeneration (\%)} \\
L  & V  &                emergent & structured &                 emergent & random & shuffled & structured \\
\midrule
3  & 6  &                 100 &    100 &                  100 &  100 &  100 &    0 \\
   & 13 &                 100 &    100 &                  100 &  100 &  100 &    0 \\
   & 27 &                 100 &    100 &                  100  &  100  & 100  &    0 \\
5  & 6  &                 100 &    100 &                  100 &  100 &  100 &    0 \\
   & 13 &                 100 &    100 &                  100  &  100  &  100 &    0 \\
   & 27 &                 100 &    100 &                  100  &  100  &  100  &    0 \\
10 & 6  &                 100 &    100 &                  98$\pm$2  &  0   &  100 &    0 \\
   & 13 &                 96$\pm$3  &    100 &                  12$\pm$10  &  0   &  2   &    0 \\
   & 27 &                 92$\pm$3  &    100 &                  0   &  0   &  0   &    0 \\
\bottomrule
\end{tabular}
\caption{DIORA-BMM}
\end{subtable}
}
\caption{Average evaluation and overgeneration coverage for the induced grammars. Standard deviations of $<0.5$ for the emergent languages are left out.}
\label{tab:coverages}
\end{table*}

\begin{table*}
  \begin{subtable}[h]{\textwidth}
    \centering
    {\footnotesize
\begin{tabular}{ll|rrrr|rrrr}
\toprule
   & {} & \multicolumn{4}{l|}{number of preterminals} & \multicolumn{4}{l}{number of terminals} \\
L   & V  &                   emergent &  random &  shuffled & structured &                emergent &  random &  shuffled & structured \\
\midrule
3  & 6  &               1.0$\pm$0.0 &   1.0 &   1.0 &    3.0 &            6.0$\pm$0.0 &   6.0 &   6.0 &    6.0 \\
   & 13 &               1.0$\pm$0.0 &   1.0 &   1.0 &    3.0 &           13.0$\pm$0.0 &  13.0 &  13.0 &   13.0 \\
   & 27 &               1.0$\pm$0.0 &   1.0 &   1.0 &    3.0 &           22.7$\pm$1.2 &  22.0 &  21.0 &   27.0 \\
5  & 6  &               2.0$\pm$0.0 &   1.0 &   2.0 &    4.0 &            6.0$\pm$0.0 &   6.0 &   6.0 &    6.0 \\
   & 13 &               3.3$\pm$1.7 &   1.0 &   2.0 &    5.0 &           12.3$\pm$0.5 &  12.0 &  13.0 &   13.0 \\
   & 27 &               2.0$\pm$1.4 &   1.0 &   1.0 &    6.0 &           20.0$\pm$2.2 &  20.0 &  21.0 &   27.0 \\
10 & 6  &               6.0$\pm$0.0 &   6.0 &   6.0 &    4.0 &            6.0$\pm$0.0 &   6.0 &   6.0 &    6.0 \\
   & 13 &              12.7$\pm$0.5 &  13.0 &  12.0 &   10.0 &           13.0$\pm$0.0 &  13.0 &  13.0 &   13.0 \\
   & 27 &              19.7$\pm$2.1 &  21.0 &  18.0 &   12.0 &           21.0$\pm$2.2 &  21.0 &  18.0 &   27.0 \\
\bottomrule
\end{tabular}
}
\caption{CCL-BMM}
\end{subtable}
\begin{subtable}[h]{\textwidth}
  \centering
  {\footnotesize
\begin{tabular}{ll|rrrr|rrrr}
\toprule
   & {} & \multicolumn{4}{l|}{number of preterminals} & \multicolumn{4}{l}{number of terminals} \\
L   & V  &                   emergent &  random &  shuffled & structured &                emergent &  random &  shuffled & structured \\
\midrule
3  & 6  &               1.0$\pm$0.0 &   1.0 &   1.0 &    3.0 &            6.0$\pm$0.0 &   6.0 &   6.0 &    6.0 \\
   & 13 &               1.7$\pm$0.5 &   1.0 &   2.0 &    3.0 &           13.0$\pm$0.0 &  13.0 &  13.0 &   13.0 \\
   & 27 &               1.3$\pm$0.5 &   1.0 &   1.0 &    3.0 &           22.7$\pm$1.2 &  22.0 &  21.0 &   27.0 \\
5  & 6  &               2.3$\pm$0.9 &   2.0 &   3.0 &    4.0 &            6.0$\pm$0.0 &   6.0 &   6.0 &    6.0 \\
   & 13 &               2.7$\pm$0.5 &   2.0 &   3.0 &    5.0 &           12.3$\pm$0.5 &  12.0 &  13.0 &   13.0 \\
   & 27 &               4.0$\pm$0.8 &   1.0 &   4.0 &    6.0 &           20.0$\pm$2.2 &  20.0 &  21.0 &   27.0 \\
10 & 6  &               6.0$\pm$0.0 &   6.0 &   6.0 &    4.0 &            6.0$\pm$0.0 &   6.0 &   6.0 &    6.0 \\
   & 13 &              13.0$\pm$0.0 &  13.0 &  13.0 &   10.0 &           13.0$\pm$0.0 &  13.0 &  13.0 &   13.0 \\
   & 27 &              20.0$\pm$1.6 &  21.0 &  18.0 &   17.0 &           21.0$\pm$2.2 &  21.0 &  18.0 &   27.0 \\
\bottomrule
\end{tabular}
}
\caption{DIORA-BMM}
\end{subtable}
\caption{Average number of pre-terminals and terminals per grammar.}
\label{tab:number_preterminals}
\end{table*}

\begin{table*}
  \begin{subtable}[h]{\textwidth}
    \centering
{\footnotesize
\begin{tabular}{ll|llll|llll}
\toprule
   & {} & \multicolumn{4}{l}{number of pre-terminal group-generating non-terminals} & \multicolumn{4}{l}{number of pre-terminal groups} \\
L   & V  &               emergent &   random &   shuffled & structured &             emergent &   random &   shuffled & structured \\
\midrule
3  & 6  &           1.0$\pm$0.0 &    1.0 &    1.0 &    2.0* &         1.0$\pm$0.0 &    1.0 &    1.0 &    3.0* \\
   & 13 &           1.3$\pm$0.5 &    1.0 &    1.0 &    1.0 &         1.3$\pm$0.5 &    1.0 &    1.0 &    1.0 \\
   & 27 &           2.0$\pm$0.0 &    1.0* &    2.0 &    1.0* &         2.0$\pm$0.0 &    1.0* &    2.0 &    1.0* \\
5  & 6  &           1.0$\pm$0.0 &    1.0 &    1.0 &    1.0 &         4.0$\pm$0.0 &    1.0* &    4.0 &    2.0* \\
   & 13 &           7.3$\pm$1.2 &    4.0 &    8.0 &    2.0* &       24.3$\pm$16.3 &    4.0 &   10.0 &    2.0 \\
   & 27 &           7.3$\pm$2.1 &    1.0* &    6.0 &    3.0 &       14.7$\pm$15.1 &    1.0 &    4.0 &    3.0 \\
10 & 6  &          14.0$\pm$6.4 &   36.0* &    1.0 &    1.0 &        34.7$\pm$0.9 &   36.0 &   36.0 &    2.0* \\
   & 13 &          46.3$\pm$9.7 &  169.0* &   16.0* &    2.0* &        78.7$\pm$6.0 &  169.0* &  137.0* &    2.0* \\
   & 27 &         64.0$\pm$15.0 &  441.0* &  233.0* &    2.0* &      192.3$\pm$64.8 &  441.0* &  262.0 &    2.0 \\
\bottomrule
\end{tabular}
  }
  \caption{CCL-BMM}
\end{subtable}
\begin{subtable}[h]{\textwidth}
  \centering
  {\footnotesize
\begin{tabular}{ll|llll|llll}
\toprule
   & {} & \multicolumn{4}{l}{number of pre-terminal group-generating non-terminals} & \multicolumn{4}{l}{number of pre-terminal groups} \\
 L  & V  &               emergent &   random &   shuffled & structured &             emergent &   random &   shuffled & structured \\
\midrule
3  & 6  &           2.0$\pm$0.8 &    2.0 &    1.0 &    3.0 &         1.0$\pm$0.0 &    1.0 &    1.0 &    3.0* \\
   & 13 &           3.0$\pm$0.0 &    3.0 &    3.0 &    3.0 &         3.0$\pm$1.4 &    1.0 &    4.0 &    3.0 \\
   & 27 &           3.0$\pm$0.8 &    2.0 &    3.0 &    4.0 &         2.0$\pm$1.4 &    1.0 &    1.0 &    3.0 \\
5  & 6  &           1.0$\pm$0.0 &    1.0 &    1.0 &    1.0 &         6.3$\pm$3.8 &    4.0 &    9.0 &    2.0 \\
   & 13 &           2.3$\pm$0.5 &    3.0 &    2.0 &    4.0* &         7.3$\pm$2.4 &    4.0 &    9.0 &    4.0 \\
   & 27 &           3.7$\pm$1.2 &   15.0* &   17.0* &    4.0 &        16.7$\pm$6.5 &    1.0 &   16.0 &    6.0 \\
10 & 6  &          11.3$\pm$6.3 &   36.0* &    3.0 &    2.0 &        35.3$\pm$0.5 &   36.0 &   36.0 &    2.0* \\
   & 13 &          71.0$\pm$9.2 &  169.0* &  117.0* &    3.0* &       130.7$\pm$9.0 &  169.0* &  144.0 &    3.0* \\
   & 27 &         96.0$\pm$14.2 &  437.0* &  262.0* &   13.0* &      225.3$\pm$21.8 &  437.0* &  265.0 &   15.0* \\
\bottomrule
\end{tabular}
  }
  \caption{DIORA-BMM}
\end{subtable}
\caption{Average number of pre-terminal groups and their generating non-terminals. The right-hand side of a production rule leading only to pre-terminals or symbols, constitutes a \emph{pre-terminal group}, while the non-terminal on the left-hand side is the respective \emph{pre-terminal group-generating non-terminals}. We indicate significant differences with the baseline value at $p<.05$ with an asterisk (*).}
\label{tab:preterminal_groups}
  \end{table*}

\begin{table*}%
  {\footnotesize
  \begin{subtable}[h]{0.5\textwidth}
    \centering

\begin{tabular}{ll|llll}
\toprule
   & {} & \multicolumn{4}{l}{average \# terminals/pre-terminal} \\
   &  &                             emergent &  random &  shuffled & structured \\
L & V &                                  &       &       &        \\
\midrule
3  & 6  &                         6.0$\pm$0.0 &   6.0 &   6.0 &    2.0* \\
   & 13 &                        13.0$\pm$0.0 &  13.0 &  13.0 &    4.3* \\
   & 27 &                        22.7$\pm$1.2 &  22.0 &  21.0 &    9.0* \\
5  & 6  &                         3.0$\pm$0.0 &   6.0* &   3.0 &    1.5* \\
   & 13 &                         6.1$\pm$4.9 &  12.0 &   6.5 &    2.6 \\
   & 27 &                        15.8$\pm$8.1 &  20.0 &  21.0 &    4.5 \\
10 & 6  &                         1.0$\pm$0.0 &   1.0 &   1.0 &    1.5* \\
   & 13 &                         1.0$\pm$0.0 &   1.0 &   1.1 &    1.3* \\
   & 27 &                         1.1$\pm$0.0 &   1.0 &   1.0 &    2.2* \\
\bottomrule
\end{tabular}

\caption{CCL-BMM}

\end{subtable}
\hfill
\begin{subtable}[h]{0.5\textwidth}
  \centering
\begin{tabular}{ll|llll}
\toprule
   & {} & \multicolumn{4}{l}{average \# terminals/pre-terminal} \\
   &  &                             emergent &  random &  shuffled & structured \\
L & V &                                  &       &       &        \\
\midrule
3  & 6  &                         6.0$\pm$0.0 &   6.0 &   6.0 &    2.0* \\
   & 13 &                         8.7$\pm$3.1 &  13.0 &   6.5 &    4.3 \\
   & 27 &                        18.7$\pm$4.8 &  22.0 &  21.0 &    9.0 \\
5  & 6  &                         3.3$\pm$1.9 &   3.0 &   2.0 &    1.5 \\
   & 13 &                         4.8$\pm$0.9 &   6.0 &   4.3 &    2.6 \\
   & 27 &                         5.3$\pm$1.5 &  20.0* &   5.2 &    4.5 \\
10 & 6  &                         1.0$\pm$0.0 &   1.0 &   1.0 &    1.5* \\
   & 13 &                         1.0$\pm$0.0 &   1.0 &   1.0 &    1.3* \\
   & 27 &                         1.0$\pm$0.0 &   1.0 &   1.0 &    1.6* \\
\bottomrule
\end{tabular}

\caption{DIORA-BMM}

\end{subtable}
}
\caption{Average number of terminals per pre-terminals. We indicate significant differences with the baseline value at $p<.05$ with an asterisk (*).}
\label{tab:wordclass}
\end{table*}

\end{document}